\documentclass{article}

\usepackage{arxiv}

\usepackage[utf8]{inputenc} 
\usepackage[T1]{fontenc}    
\usepackage{hyperref}       
\usepackage{url}            
\usepackage{booktabs}       
\usepackage{amsfonts}       
\usepackage{nicefrac}       
\usepackage{microtype}      
\usepackage{lipsum}
\usepackage{svg}
\usepackage{graphicx}
\graphicspath{ {./images/} }
\usepackage{amsmath} 
\usepackage{pdflscape}
\usepackage{xcolor}
\usepackage[normalem]{ulem}
\usepackage{comment}
\usepackage{longtable}
\usepackage{tabularx}
\usepackage{caption} 
\usepackage{subcaption}
\usepackage{longtable}
\usepackage{booktabs}
\usepackage[utf8]{inputenc}
\usepackage{changepage}

\usepackage{array}
\usepackage{float}
\newlength{\extralength}
\newlength{\fulllength}
\newcolumntype{C}{>{\centering\arraybackslash}X}

\title{Comparing YOLOv5 Variants for Vehicle Detection: A Performance Analysis}

\author{Athulya Sundaresan Geetha \textsuperscript{*}\\[1ex]
\begin{minipage}[t]{0.90\textwidth}
\centering
\scriptsize Department of Computer Science, Huddersfield University, Queensgate, Huddersfield HD1 3DH, UK; \\
\textsuperscript{*}Correspondence: U2282847@unimail.hud.ac.uk;
\end{minipage}}

\begin{document}

\maketitle
\begin{abstract}Vehicle detection is an important task in the management of traffic and automatic vehicles. This study provides a comparative analysis of five YOLOv5 variants, YOLOv5n6s, YOLOv5s6s, YOLOv5m6s, YOLOv5l6s, and YOLOv5x6s, for vehicle detection in various environments. The research focuses on evaluating the effectiveness of these models in detecting different types of vehicles, such as Car, Bus, Truck, Bicycle, and Motorcycle, under varying conditions including lighting, occlusion, and weather. Performance metrics such as precision, recall, F1-score, and mean Average Precision are utilized to assess the accuracy and reliability of each model. YOLOv5n6s demonstrated a strong balance between precision and recall, particularly in detecting Cars. YOLOv5s6s and YOLOv5m6s showed improvements in recall, enhancing their ability to detect all relevant objects. YOLOv5l6s, with its larger capacity, provided robust performance, especially in detecting Cars, but not good with identifying Motorcycles and Bicycles. YOLOv5x6s was effective in recognizing Buses and Cars but faced challenges with Motorcycle class. 
\end{abstract}


\section{Introduction}
In the field of autonomous driving and traffic management, accurately detecting vehicles is essential for ensuring safety and efficiency. Despite advancements in computer vision, detecting various types of vehicles in diverse environments presents several challenges. This study focuses on the detection of multiple vehicle types using the YOLOv5 models. Variations in lighting conditions, such as sunlight glare or shadows, can affect vehicle visibility in images. The presence of other objects and varying road conditions can obstruct the view of vehicles, leading to partial or complete occlusions.  Low-resolution or blurred images due to dynamic vehicle movement or poor camera quality also reduce the effectiveness of detection.

Histogram of oriented gradients (HOG) extracted horizontal and vertical features, whereas vertical histograms oriented gradients (VHOG) extracted vertical ones \cite{RN1}. Support vector machine (SVM) and extreme learning machine (ELM) classified them but could not identify them during dynamic movements. SVM, a machine learning technique, was unable to identify movements despite using localized images and features during dynamic movements from spatiotemporal areas for classification \cite{RN2, RN3, RN4}. CNNs, including Faster R-CNN and GoogleNet, improved detection \cite{RN5}. Models like MobileNet, MaskR-CNN, and PoseNet enhanced performance and segmentation.  Additional models, such as AlexNet, GoogleNet, ResNet, VGG-Net, R-CNN, Fast-RCNN, and Faster-RCNN, further advanced object identification \cite{RN6, RN7, RN8, RN9, RN10, RN11, RN12}.

As challenges in data handling and real-time execution persist for two-stage detection processes, YOLO (You Only Look Once) models were introduced to overcome these problems. YOLOv1 and Fast YOLO featured 24 and 9 layers, respectively \cite{RN14}. YOLOv2, based on DarkNet-19, enhanced feature extraction and dimensional accuracy \cite{RN15}. YOLOv3 using DarkNet-53 integrated residual networks and binary cross-entropy \cite{RN16}, while YOLOv4 employed CSPDarkNet53 and PANet for improved performance \cite{RN17}.

YOLOv5 enhanced speed with the CSP-PAN neck and SPPF head \cite{RN18}. YOLOv6 reduced the computational complexity \cite{RN19}, whereas while YOLOv7 improved detection accuracy with additional heads \cite{RN20}. YOLOv8 supports object identification and segmentation with a modified CSPDarkNet53 and PAN-FPN neck. YOLOv9 and YOLOv10, featuring Generalized Efficient Layer Aggregation and Programmable Gradient Information, improved detection accuracy, with YOLOv10 also removing non-maximum suppression \cite{RN21, RN22}.

The aim of this research is to look in detail at various yolov5 variants and present comparative analysis of the performance of YOLOv5 variants in the identification of vehicles based on a given dataset. Performance metrics including accuracy, recall, mAP50, mAP50-95, F1 score, and confusion matrix are presented to find the best model among the three. This can be helpful in real-time traffic management systems on a national and internal scale. know when the knife is incorrectly used majorly based on the safety guidelines to prevent mishaps.

The rest of the paper starts with a review of present detection technologies. This is followed by the Methodology section. We then present the results for the various variants before moving onto the discussion. The conclusion presents a summary of the research and future direction.

\section{Literature Review} 

In computer vision, accurate detection and tracking have grown to be an essential part of a wide range of applications, such as traffic management and autonomous driving. Advancements in technology have transitioned from simple algorithms for the detection of object to advanced deep learning techniques enabling complex vehicle recognition tasks across diverse environments. In this section, the initial vehicle detection to different variants of YOLO versions are examined, highlighting the importance of contributions and innovations influencing modern vehicle monitoring systems.

Using bounding boxes, the identification and adaptation of the vehicles are known as vehicle detection, whereas the prediction of vehicle movements through their trajectories is tracking \cite{RN18}. Previous algorithms primarily focused on eliminating background and extracting user-defined features but had challenges like dynamic backgrounds and different climatic conditions \cite{RN19}. Hence, Barth et al. introduced a method named Stixels that leverages a color schema to provide movement details \cite{RN20}. Convolutional neural networks (CNNs) have been employed in order to avoid hidden information, background changes, and delays, resulting in improved detection accuracy \cite{RN4}. RCNN, FRCNN, SSD, and ResNet are the CNN architectures that have been the subject of numerous studies \cite{RN5, RN21,RN22, RN23, alif2024boltvision, alif2024lightweight, alif2024attention, alif2024state}. Furthermore, Azimjonov and Özmen conducted a comparative analysis of traditional machine learning and deep learning algorithms for road vehicle detection \cite{RN8}. YOLO for detection and CNN for tracking have demonstrated superior performance when compared to other machine learning models.

YOLO has revolutionized vehicle detection by treating it as a regression task, accurately identifying vehicle locations, types, and confidence scores, and improves detection speed to handle blur images by generating bounding boxes and class probabilities. YOLOv2 utilized GPU capabilities and the Anchor Box technique to enhance vehicle detection, classification, and tracking \cite{RN24}. YOLOv3 was trained on the detection of five classes, cars, trucks, street signs, people, and traffic lights, in various weather conditions \cite{RN25}. YOLOv4 was implemented to increase the speed at which slow-moving cars in video streams could be detected \cite{RN26}.

With YOLOv5's infrared camera and effective framework, it recognized large cars in snowy conditions, allowing real-time parking slot predictions. YOLO-v5n is highly effective due to its exceptional balance of high accuracy and minimal computational requirements, achieving 96.8\% mAP@0.5 while being lightweight for efficient edge deployment \cite{hussain2023yolo}. YOLOv6 improved upon previous network architecture and training methods to achieve greater real-time detection accuracy \cite{RN28}. To improve tracking and decision-making in urban environments, YOLOv7 was designed with the purpose of detecting, tracing, and measuring vehicles on highways \cite{RN29, RN30}.

YOLOv8 was enhanced with a Spatial Attention Module (SAM) and Spatial Pyramid Pooling (SPP) module to improve feature representation and capture vehicle features at multiple scales, respectively \cite{RN31}. An industrial fall detection system using YOLOv8 variants was developed, finding YOLOv8m optimal for balancing accuracy (mAP 0.828) and computational efficiency, outperforming YOLOv8l \cite{pereira2024fall}. YOLOv8 achieved an accuracy of 95.6\% and 94.6\% in vehicle detection using advanced feature extraction, Scale Invariant Feature Transform, Oriented FAST and Rotated BRIEF, and KAZE and a Deep Belief Network classifier \cite{RN32}. Because of YOLOv10's Non-Maximum Suppression-free design, which lowers latency and speeds up post-processing, it overlooks smaller objects as it has fewer parameters and lower confidence scores \cite{RN33}. A study comparing YOLOv5, YOLOv8, and YOLOv10 for knife safety found that YOLOv5 was most effective at detecting dynamic movement, i.e., hand-blade contact, while YOLOv8 excelled at identifying curled fingers due to its enhanced ability to recognize fine details \cite{geetha2024comparative}. When comparing YOLOv8 and YOLOv10 for vehicle detection, it is noted that YOLOv10 was generally better at detecting smaller (bicycles) and more complex (truck) vehicles and YOLOv8 performed well in detecting cars \cite{sundaresan2024comparative}.

\section{Methodology}
\subsection{Dataset}
\begin{figure}[H]
\begin{adjustwidth}{-\extralength}{0cm}
\centering
\includegraphics[width=15cm]{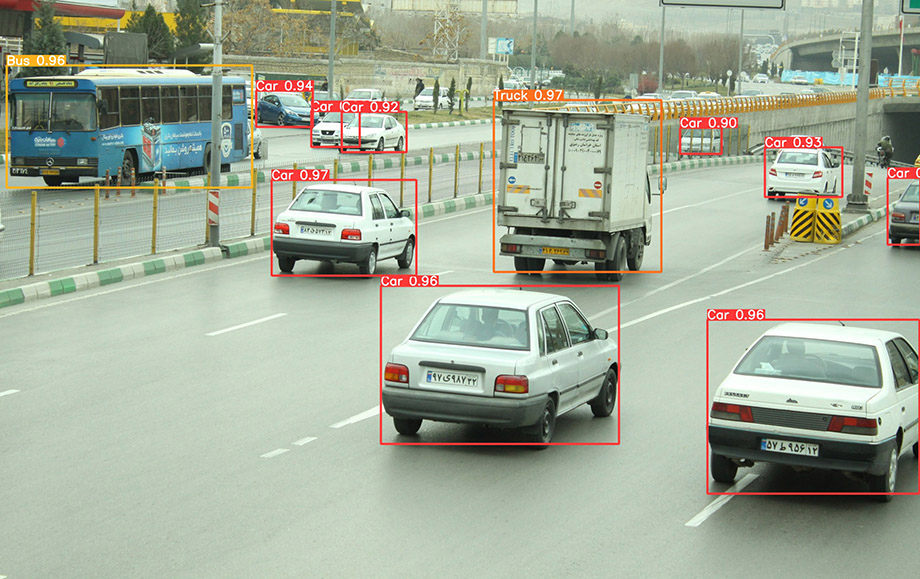}
\end{adjustwidth}
\caption{Sample image of the vehicle dataset \cite{dataset}.}
\label{Figure:1}
\end{figure}

The dataset utilized in this study was sourced from the repository ‘GitHub - MaryamBoneh/Vehicle-Detection: Vehicle Detection Using Deep Learning and YOLO Algorithm’ \cite{dataset}. It includes 1321 images, containing 1196 images for training, 8 for testing, and 117 for validation. With a resolution of 416 × 416 pixels, an image is labelled as five classes: Car, Motorcycle, Truck, Bus, and Bicycle. An example image from the dataset is shown in Figure \ref{Figure:1}.

\subsection{Data Augmentation}
Data augmentation is a crucial task to expand the dataset by generating additional images with original images. This enables the model to learn from a broader range of scenarios, improving its ability to generalize to unseen data and reducing the risk of overfitting. Data augmentation is particularly useful for addressing class imbalances by creating more samples of underrepresented categories, thus ensuring a more balanced and unbiased model. Introducing noise during augmentation helps the model adapt to real-world variations. Since collecting large amounts of real-time data is often impractical, augmentation provides a viable solution by significantly increasing the dataset size.

Challenges including varying light conditions, weather changes, occlusions, and different angles complicate vehicle detection and classification. Data augmentation helps to resolve these challenges by diversifying the training dataset, introducing variations such as changes in lighting, angles, and occlusions. Utilizing augmented data allows models to improve their detection and classification accuracy across various conditions, thereby enhancing both their performance and robustness.

\subsubsection{Random Crop}
Cropping enhances the model's performance to detect vehicles in scenarios such as the object not completely visible, scaled uniformly, or appear at different angles from the camera. In this case, the images are cropped with a minimum zoom of 0\%, retaining the original image dimensions (416 × 416 pixels), to an extent of a maximum zoom of 20\% (332 × 332 pixels), where 20\% of the image width is cropped from either side. This method improves effectiveness in real-world scenarios where vehicles may vary in position, size, and shape. Equation \ref{eq:1} is calculated to crop 20\% from each side of an image.

\begin{equation}
    S^{\prime}=S[ top +p \times height, bottom -p \times height, left +p \times width, right -p
    \label{eq:1}
\end{equation}

\ \(S^{\prime}\) is the output image; \(S\) is the input image; and \(p\) indicates cropping percentage.

\subsubsection{Random Rotation}
The image is rotated between -15 and +15 degrees, altering the position of vehicles within the frame (Equation \ref{eq:2}). The bounding box must be adjusted to accommodate the vehicle as seen from a different perspective. When dealing with rotating cameras, vehicles may not appear aligned in the frames and could be viewed from different angles, thus improving model learning.

\begin{equation}
    S^\prime\ =\ rotate(S,\theta)
    \label{eq:2}
\end{equation}

 The 2-dimensional rotational transformation of an image is given in Equation \ref{eq:3}.

\begin{equation}
    R\left(\theta\right)=\left[\begin{matrix}cos(\theta)&-sin(\theta)\\sin(\theta)&cos(\theta)\\\end{matrix}\right]
    \label{eq:3}
\end{equation}

\ \( R(\theta) \) is rotation matrix.

\subsubsection{Random Shear}
This technique involves slanting the image along the x or y axis, causing a tilt that distorts objects within the image. For vehicle detection, applying random shearing at ±10 degrees simulates various angles and creates geometric distortions (Equation \ref{eq:4}). This helps the model adapt to different perspectives and distortions, improving its effectiveness in real-world scenarios where vehicles may appear at various angles and positions.

\begin{equation}
    S^\prime\ =\ S\cdot S
    \label{eq:4}
\end{equation}

Shear matrix is denoted by \textit{s}.

\subsubsection{Random Grayscale}
Grayscaling 15\% of colored vehicle images emphasizes shapes and textures over color. This technique introduces variations such as different lighting conditions and monochromatic scenarios, improving the model’s ability to detect vehicles under diverse settings (Equation \ref{eq:5}). By converting a portion of images to grayscale, the model focuses on structural features, which aids in accurate vehicle detection even with color variations.

\begin{equation}
    S' = 
    \begin{cases} 
        0.299 \times R + 0.587 \times G + 0.114 \times B, & \text{with probability } p \\
        S, & \text{with probability } (1-p)
    \end{cases}
    \label{eq:5}
\end{equation}

\(R\), \(G\), and \(B\) indicate red, green, and blue, respectively; and \(p\) represents the transformation probability.

\subsubsection{Saturation}
Adjusting saturation from -25\% to +25\% introduces various color intensities and lighting conditions to vehicle images (Equation \ref{eq:7}). This variation in saturation enhances the model's ability to recognize vehicles under different color and light settings. The saturation adjustment factor ($\alpha$), ranging from -25\% to +25\%, helps the model learn to handle diverse color intensities effectively.

\begin{equation}
    S^\prime\ =\text{adjust\_saturation}(S,\alpha)
    \label{eq:7}
\end{equation}

\subsubsection{Brightness}
Brightness augmentation is crucial for enabling the model to handle varying lighting conditions encountered in different weather scenarios. In this study, brightness levels of vehicle images were adjusted between -15\% and +15\% (Equation \ref{eq:8}). This adjustment exposes the images to a range of lighting conditions, helping the model adapt to different brightness levels for accurate vehicle detection in diverse environments.

\begin{equation}
    S^\prime\ =\ clip(S\ +\ \beta\ \times255,\ 0,255)
    \label{eq:8}
\end{equation}

\ \(\beta\) is brightness adjustment factor.

\subsubsection{Blur}
Blurring introduces a blur effect to vehicle images to simulate real-world conditions where images may lack sharpness and contrast. This technique helps the model learn to detect vehicles even when objects are out of focus, which is crucial for identifying vehicles in various scenarios. By reducing sharpness and contrast, blurring also decreases noise and enhances feature extraction. The Gaussian blur kernel's standard deviation, represented by $\sigma$ (Equation \ref{eq:9}), varies up to 2.5 pixels. This approach aids in vehicle detection by improving accuracy through reduced noise and more effective feature extraction.

\begin{equation}
    S^\prime\ =\ Blur(S,\sigma),
    \label{eq:9}
\end{equation}

\subsubsection{Random Noise}
Adding noise to vehicle images aids the model in recognizing important features despite distortions (Equation \ref{eq:10}). This process introduces noise to 0.1\% of the pixels, creating random black and white spots in various areas of the image. By intentionally altering pixel values, this technique simulates real-world conditions where images may contain unexpected distortions. The noise (N) applied to the original image helps improve the model’s accuracy in vehicle detection under less ideal conditions.

\begin{equation}
    S^\prime\ =\ S+N
    \label{eq:10}
\end{equation}

\subsubsection{HSV}
The Hue, Saturation, and Value (HSV) model is used to control image brightness and color properties. In this context, hue defines the color, saturation adjusts the color intensity, and value sets the brightness level. To reduce the brightness to 40\%, the value is set to 0.4. The hue is adjusted to 0.015 to shift the color tone slightly, while the saturation is set at 0.7 to increase the color intensity to 70\%. The brightness adjustment for each pixel is represented by $\alpha$ (Equation \ref{eq:11}).

\begin{equation}
    S^\prime\ =\ \alpha S
    \label{eq:11}
\end{equation}

\subsubsection{Translate}
Translation adjusts the position of an image along the horizontal and vertical axes. By setting the translation factor to 0.1, the image is shifted 10

\begin{equation}
    S^\prime\left(x+t_x\times w i d t h,\ y+t_y\ \times h e i g h t\right)=S(x,\ y)
    \label{eq:12}
\end{equation}

\subsubsection{Mosaic}
Mosaic augmentation creates a single image by combining four different images, each contributing a section of the final composite. When the mosaic parameter is set to 1.0, this technique is applied to the entire training set, meaning every image is processed in this way to increase the diversity and complexity of the training data. The coordinates of the resulting mosaic image are expressed as $(x^\prime)$ and $(y^\prime)$, while the original image coordinates are labeled as \( x_i \) and \( y_i \). The intensity of each image section is represented by \( S_i \). This method is useful in vehicle detection for generating varied training scenarios (Equation \ref{eq:13}).

\begin{equation}
    S^\prime\left(x^\prime,\ y^\prime\right)=\ S_i(x_i,\ y_i)
    \label{eq:13}
\end{equation}

\subsubsection{Random Erasing}
The erasing technique involves removing or obscuring portions of an image to create occlusions. In this context, setting erasing=0.4 means that 40\% of the image is intentionally hidden or removed. After applying this erasing method, the pixel intensity in the altered image is represented by \( D^\prime(x,\ y) \). This approach is useful in vehicle detection to simulate real-world scenarios where vehicles might be partially obstructed (Equation \ref{eq:14}).

\begin{equation}
   S^\prime(x,\ y)=\left\{\begin{matrix}0,&if\ \left(x,\ y\right)is\ within\ the\ erased\ region\\S\left(x,\ y\right),&otherwise\\\end{matrix}\right.
    \label{eq:14}
\end{equation}

Data augmentation is essential for vehicle detection as it helps generate extensive and varied datasets, which improves the model’s ability to generalize and prevents overfitting. By creating diverse scenarios through augmentation, the model can learn to identify vehicles under different conditions and variations, enhancing its overall accuracy and robustness.

\subsection{YOLOv5 Architecture}

The YOLOv5 model architecture is structured into three main components: the backbone, neck, and head. The backbone employs a ResNet-based Cross-Stage Partial (CSP) network to enhance efficiency with cross-stage connections and incorporates several Spatial Pyramid Pooling (SPP) blocks to extract diverse features while reducing computational load. The neck includes a Path Aggregation Network (PAN) module and upscaling layers to improve the resolution of feature maps and optimize information flow between layers. The head of YOLOv5 consists of three convolutional layers responsible for predicting bounding boxes, class labels, and confidence scores.

Anchor-based predictions in YOLOv5 associate bounding boxes with predefined anchor boxes of specific dimensions and shapes. The model's loss function incorporates Binary Cross-Entropy to handle class and objectness losses, while location loss is computed using Complete Intersection over Union (CIoU). The method for calculating objectness loss varies based on the dimensions of the prediction layer \cite{RN13}. The YOLOv5 architecture, including details on the number of filters, filter sizes, and the repetition count of each layer, is outlined in Table \ref{tab:yolov5_architecture}.

\begin{table}[H]
\caption{YOLOv5 Architecture.\label{tab:yolov5_architecture}}
\begin{adjustwidth}{-\extralength}{0cm}
    \begin{tabularx}{\fulllength}{CCCCCCC}
        \toprule
        \textbf{Layer} & \textbf{Activation} & \textbf{Filters} & \textbf{Size} & \textbf{Repeat} & \textbf{Parameters} & \textbf{Output Size} \\
        \toprule
        Image          & -          & -       & -          & -      & -                  & 640 × 640 \\
        Conv0          & ReLU       & 16      & 3 × 3 / 2  & 1      & 448                & 320 × 320 \\
        Conv1          & ReLU       & 32      & 3 × 3 / 2  & 1      & 4,640              & 160 × 160 \\
        Conv2          & ReLU       & 64      & 3 × 3 / 2  & 1      & 18,496             & 80 × 80  \\
        Conv3          & ReLU       & 128     & 3 × 3 / 2  & 1      & 73,856             & 40 × 40  \\
        Conv4          & ReLU       & 256     & 3 × 3 / 2  & 1      & 295,168            & 20 × 20  \\
        C3             & ReLU       & 128     & 1 × 1      & 1      & 16,512             & 20 × 20  \\
        Conv5          & ReLU       & 128     & 3 × 3      & 1      & 147,584            & 20 × 20  \\
        C3             & ReLU       & 256     & 1 × 1      & 1      & 33,024             & 20 × 20  \\
        Conv6          & ReLU       & 256     & 3 × 3      & 1      & 590,080            & 20 × 20  \\
        SPP            & -          & -       & -          & -      & -                  & 20 × 20  \\
        C3             & ReLU       & 512     & 1 × 1      & 1      & 131,584            & 20 × 20  \\
        Upsample       & -          & -       & -          & 1      & -                  & 40 × 40  \\
        C3             & ReLU       & 256     & 1 × 1      & 1      & 65,536             & 40 × 40  \\
        Upsample       & -          & -       & -          & 1      & -                  & 80 × 80  \\
        C3             & ReLU       & 128     & 1 × 1      & 1      & 32,768             & 80 × 80  \\
        Conv7          & ReLU       & 128     & 3 × 3      & 1      & 147,584            & 80 × 80  \\
        C3             & ReLU       & 64      & 1 × 1      & 1      & 8,192              & 80 × 80  \\
        Conv8          & ReLU       & 64      & 3 × 3      & 1      & 36,928             & 80 × 80  \\
        C3             & ReLU       & 32      & 1 × 1      & 1      & 2,816              & 80 × 80  \\
        \bottomrule
    \end{tabularx}
\end{adjustwidth}
\end{table}

The model architecture employs a sequence of convolutional layers, utilizing ReLU activation functions, to progressively reduce the image dimensions from 640 × 640 pixels down to 320 × 320, and eventually to 20 × 20 pixels. Subsequent to this reduction, the C3 layer integrates multiple 1 × 1 and 3 × 3 convolutions in a repeated pattern, preserving the spatial dimensions of the output. The Spatial Pyramid Pooling (SPP) layer aggregates features from various scales while retaining the 20 × 20 pixel dimension. Finally, upscaling operations are applied to enlarge the feature maps, enhancing their resolution.

YOLOv5 was developed to enhance both performance and accuracy over its predecessors, incorporating advanced techniques for feature extraction, aggregation, and anchor-based prediction. To ensure compatibility with iOS devices, the model utilizes ONNX and CoreML frameworks instead of PyTorch. YOLOv5x demonstrated an average precision of 50.7\% on the MS COCO dataset and achieved impressive processing speeds of 200 frames per second with 640-pixel images. For larger images at 1534 pixels, the model's average precision improved to 55.8\%.

\section{Experimental Results}
The performance metrics of various YOLOv5 variants during both the training and validation phases in the context of vehicle detection were explored. The models were trained and validated using a high-performance PyTorch environment with NVIDIA GPUs. Throughout 40 epochs, hyperparameters were adjusted to enhance the models' performance and accuracy. With a momentum of 0.9 and a learning rate of 0.001, the AdamW optimizer was used for the training process. The group parameters were carefully adjusted so that certain weights were exempt from decay, and biases were subjected to L2 regularization with a decay rate of 0.0005. This approach helped prevent overfitting while maintaining the model's generalizability. The generalizability of the models was maintained while overfitting was avoided.

\begin{figure}[H]
\begin{adjustwidth}{-\extralength}{0cm}
\centering
\includegraphics[height=8cm]{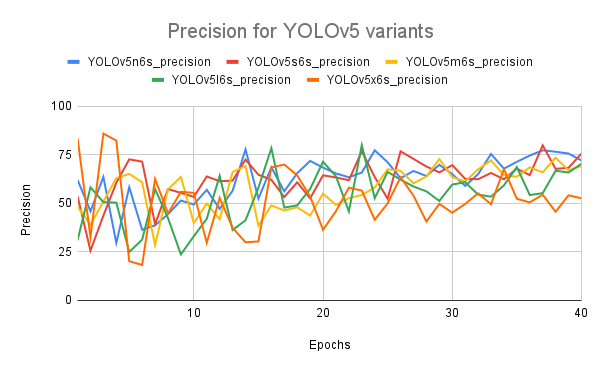}
\end{adjustwidth}
\caption{Comparison of precision values across YOLOv5 versions.}
\label{Figure:2}
\end{figure}

The precision over 40 epochs for different YOLOv5 variants is displayed in Figure \ref{Figure:2}. It is a measure of the ability of the model to identify true positives among all detected objects. High precision indicates that the model makes fewer false positive errors. Throughout the training process, the precision for all YOLOv5 versions fluctuates significantly, specifically in the earlier epochs. YOLOv5s6s and YOLOv5m6s variants show good precision, suggesting moments where the models were highly confident in their predictions. However, as training progresses, these fluctuations diminish, and precision levels begin to stabilize. This indicates that while the models are improving in reducing false positives, there is still variability in their performance across different epochs.

\begin{figure}[H]
\begin{adjustwidth}{-\extralength}{0cm}
\centering
\includegraphics[height=8cm]{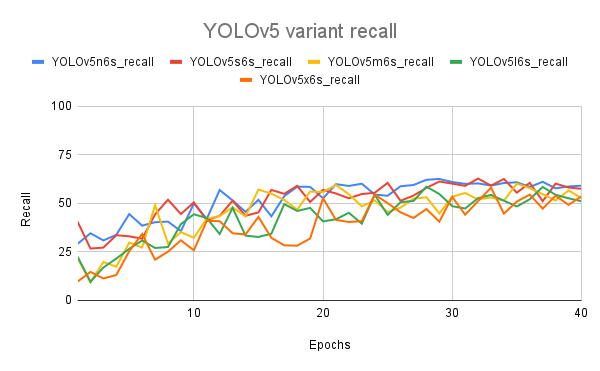}
\end{adjustwidth}
\caption{Comparison of recall values for YOLOv5 versions.}
\label{Figure:3}
\end{figure}

Figure \ref{Figure:3} illustrates the recall metric over 40 epochs for the various YOLOv5 variants. Recall measures the model's ability to detect all relevant objects in the dataset, stating it reflects the rate of true positive detections among objects. In the early processes of training, the recall starts at a lower level, indicating that the models were initially missing a significant number of objects. However, over time, the recall steadily improves across all variants, with YOLOv5s6s and YOLOv5m6s showing more upward trends. By the 40 epochs, all variants have achieved better recall rates, although YOLOv5n6s and YOLOv5l6s exhibit a slight lag. This improvement in recall suggests that the models have become more effective at detecting a higher proportion of objects in the images by the end of training, reducing the number of missed detections.

\begin{figure}[H]
\begin{adjustwidth}{-\extralength}{0cm}
\centering
\includegraphics[height=8cm]{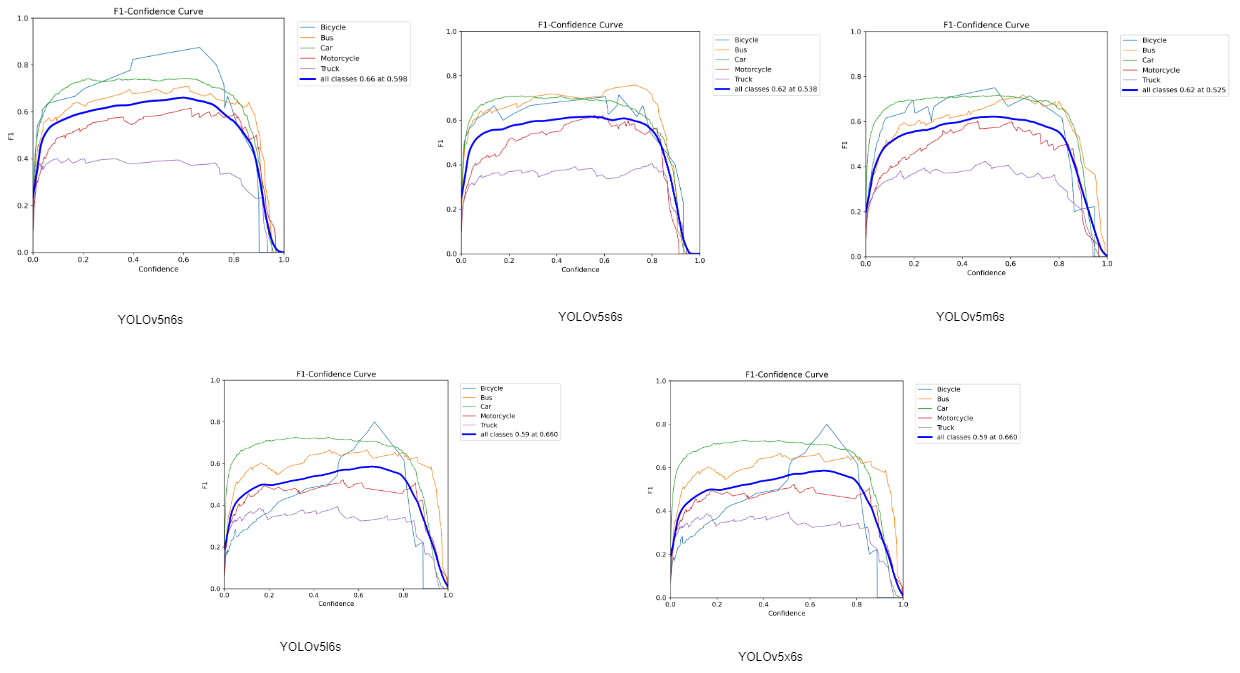}
\end{adjustwidth}
\caption{Comparison of precision values across YOLOv5 versions.}
\label{Figure:4}
\end{figure}

The F1-Confidence curve for YOLOv5n6s model shows that it reaches its peak F1 score of 0.66 at a confidence threshold of 0.598 (Figure \ref{Figure:4}). "Bicycle" and "Truck" show relatively good performance with higher F1 scores compared to "Motorcycle," which struggles with lower scores, likely indicating that the model has difficulty distinguishing motorcycles from other objects at various confidence levels.

The peak F1 score for YOLOv5s6s model is 0.62 at a confidence threshold of 0.538, slightly lower than the nano version, suggesting that even though this model is still lightweight, it may not perform as well for all classes. The individual class curves show improved performance for "Bus" and "Car" compared to the nano model, but the F1 score for "Motorcycle" remains low, indicating persistent challenges in classifying the object type accurately.

In YOLOv5m6s, the F1-Confidence curve shows a peak F1 score of 0.62 at a confidence threshold of 0.525 for all classes, similar to the small model but with slight improvements in consistency across the confidence range. The curves indicate that "Bus," "Car," and "Bicycle" classifications perform well, while Motorcycle still lags, although slightly better than in the previous model. This suggests that the medium model offers marginally better generalization across most classes.

YOLOv5l6s model achieves a peak F1 score of 0.59 at a confidence threshold of 0.66 for all classes, indicating a slight dip in overall F1 score compared to the previos models. However, individual class performance shows improvements, especially in the Bicycle and Bus classes, suggesting that the large model can better capture features relevant to these objects. The decline in overall F1 score might indicate that while it is better at certain tasks, it might be overfitting to some classes or struggling with balancing performance across all classes.

YOLOv5x6s is the most complex and highest-capacity model in the YOLOv5 family. It also reaches a peak F1 score of 0.59 at a confidence threshold of 0.66. It exhibits the highest curves for Bus and Car, indicating that it excels in these categories. However, the Motorcycle class still presents challenges. As the model size increases, there are generally improvements in performance for specific classes like Bus and Car, but challenges remain for a class like Motorcycle.

\begin{figure}[H]
\begin{adjustwidth}{-\extralength}{0cm}
\centering
\includegraphics[height=6cm]{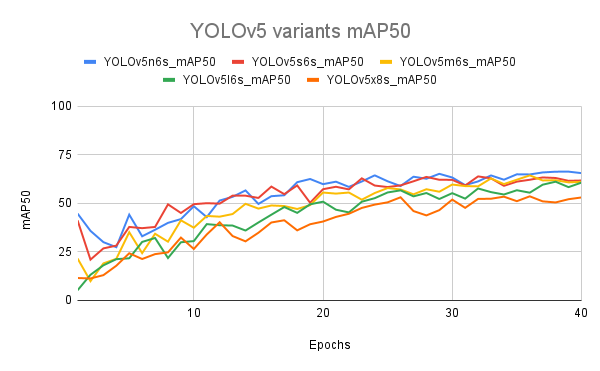}
\end{adjustwidth}
\caption{Comparison of mAP50 values across all five YOLOv5 variants.}
\label{Figure:5}
\end{figure}

The mean Average Precision (mAP) at an Intersection over Union (IoU) threshold of 0.50 (mAP50) across 40 epochs for various YOLOv5 variants is shown in Figure \ref{Figure:5}. The mAP50 is important as it measures the accuracy of object detection models in identifying objects within a 50\% overlap between the predicted bounding box and the original. In the early epochs, the YOLOv5n6s variant demonstrates a relatively strong performance, with a steeper increase in mAP50. However, as training progresses, its mAP50 levels off. The other variants, YOLOv5s6s, YOLOv5m6s, YOLOv5l6s, and YOLOv5x6s, show a more gradual improvement. While they start with slightly lower mAP50 values, they steadily converge at epoch 40.

\begin{figure}[H]
\begin{adjustwidth}{-\extralength}{0cm}
\centering
\includegraphics[height=9cm]{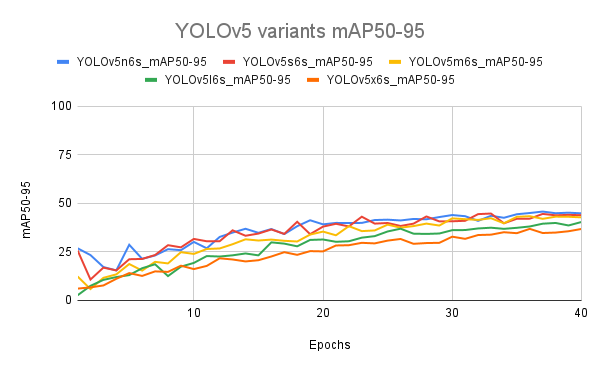}
\end{adjustwidth}
\caption{Comparison of mAP50-95 values across three models.}
\label{Figure:6}
\end{figure}

The mean Average Precision across a range of IoU thresholds from 0.50 to 0.95 (mAP50-95) over 40 epochs is illustrated in Figure \ref{Figure:6}. The mAP50-95 is more rigorous than mAP50, as it averages precision across multiple thresholds. Initially, all YOLOv5 variants exhibit an average mAP50-95, reflecting the difficulty in achieving high precision at higher IoU thresholds. In the beginning, YOLOv5n6s shows a slightly better start, but over time, the performance of all variants gradually improves. By the end of the training, YOLOv5s6s, YOLOv5m6s, and YOLOv5x6s variants exhibit superior performance. This convergence indicates that the models are refining their ability to make more accurate predictions across varying levels of overlap, although the improvements are more incremental compared to mAP50.

\begin{figure}[H]
\begin{adjustwidth}{-\extralength}{0cm}
\centering
\includegraphics[height=15cm]{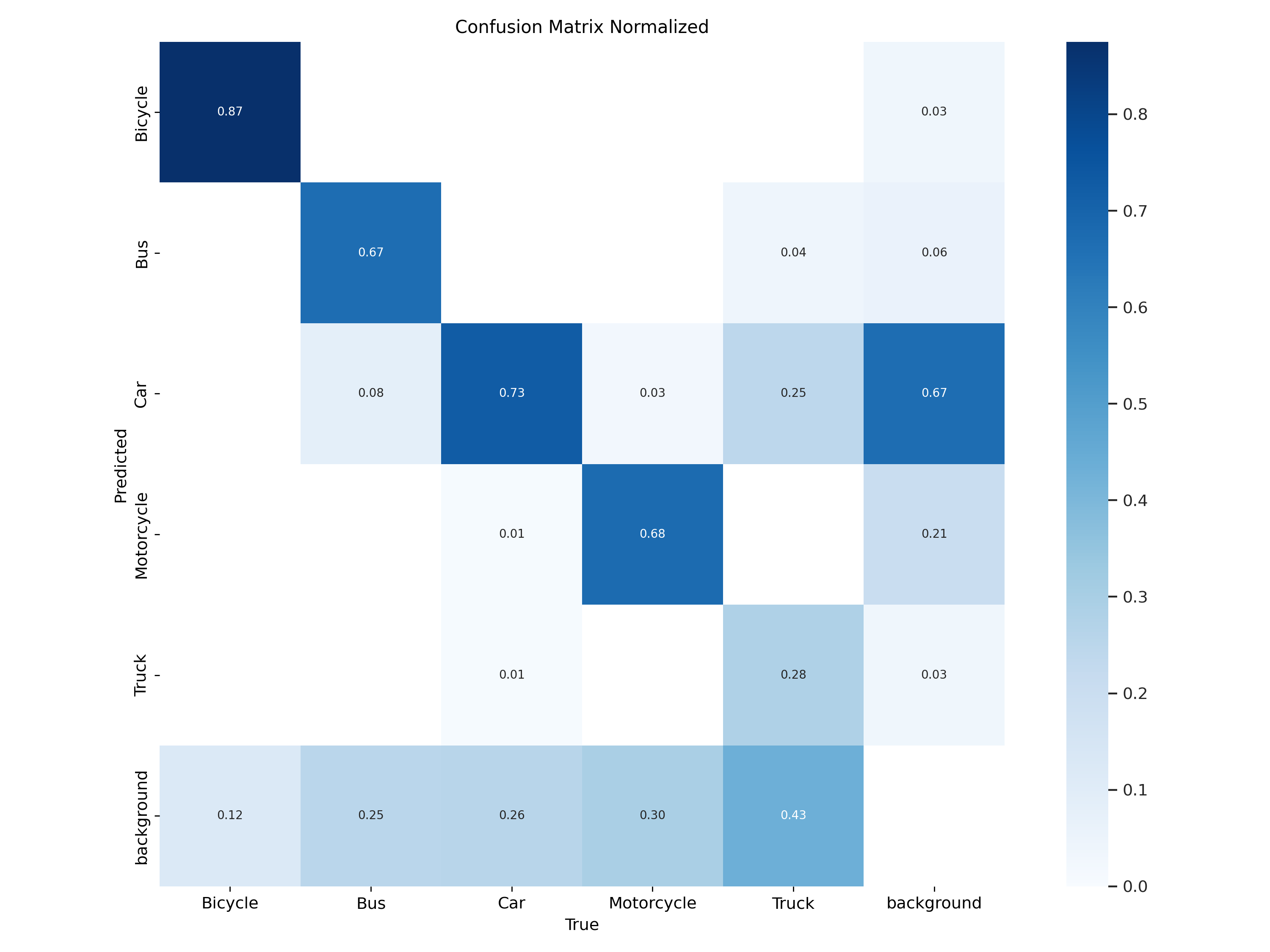}
\end{adjustwidth}
\caption{YOLOv5n6s normalized confusion matrix.}
\label{Figure:7}
\end{figure}

\begin{figure}[H]
\begin{adjustwidth}{-\extralength}{0cm}
\centering
\includegraphics[height=15cm]{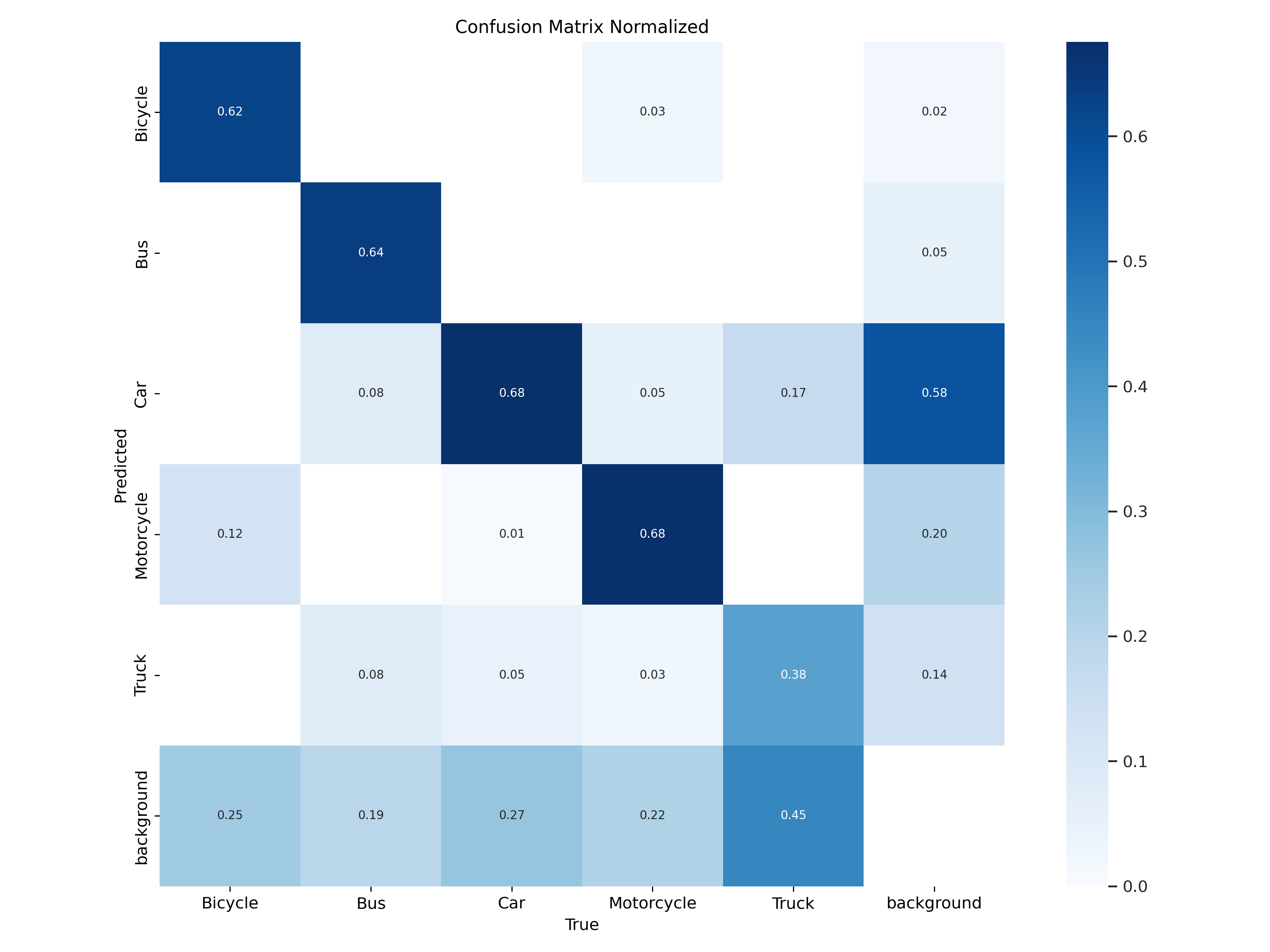}
\end{adjustwidth}
\caption{YOLOv5s6s normalized confusion matrix.}
\label{Figure:8}
\end{figure}

The normalized confusion matrix provided the classification accuracy for all the classes (Bicycle, Bus, Car, Motorcycle, and Truck) in all five model variants. YOLOv5n6s model appears to have slightly better performance in classifying "Car" but continues to struggle with distinguishing "Truck" from "Car," as indicated by a significant portion of misclassified instances (Figure \ref{Figure:7}). For the YOLOv5s6s variant, the confusion matrix highlights some weaknesses, particularly in distinguishing between Motorcycle and Bicycle, due to their similar shapes and sizes. On the other hand, Bus class is classified correctly (Figure \ref{Figure:8}). This variant appears to struggle more with smaller and more similar objects.

\begin{figure}[H]
\begin{adjustwidth}{-\extralength}{0cm}
\centering
\includegraphics[height=15cm]{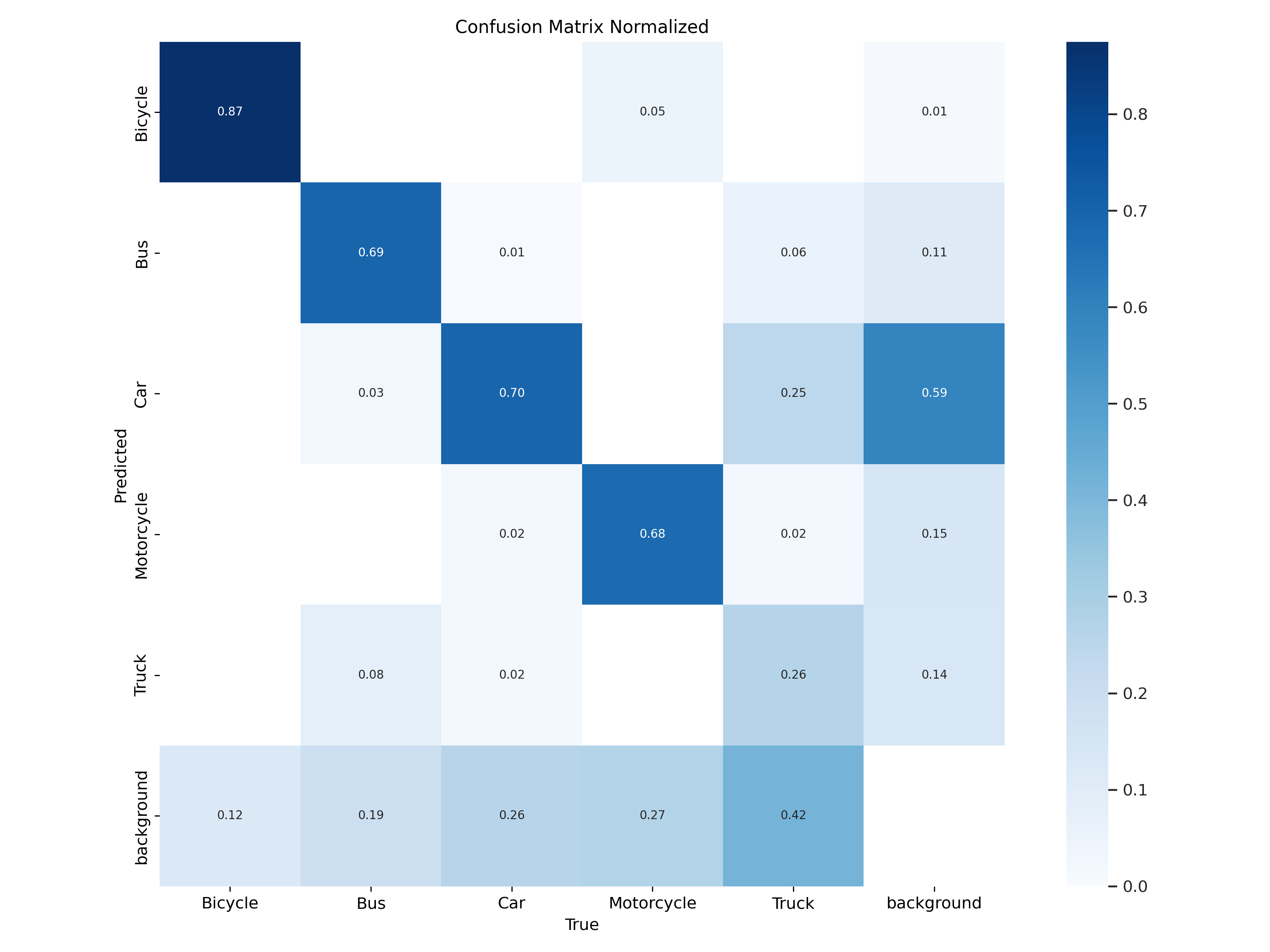}
\end{adjustwidth}
\caption{YOLOv5m6s normalized confusion matrix.}
\label{Figure:9}
\end{figure}

\begin{figure}[H]
\begin{adjustwidth}{-\extralength}{0cm}
\centering
\includegraphics[height=15cm]{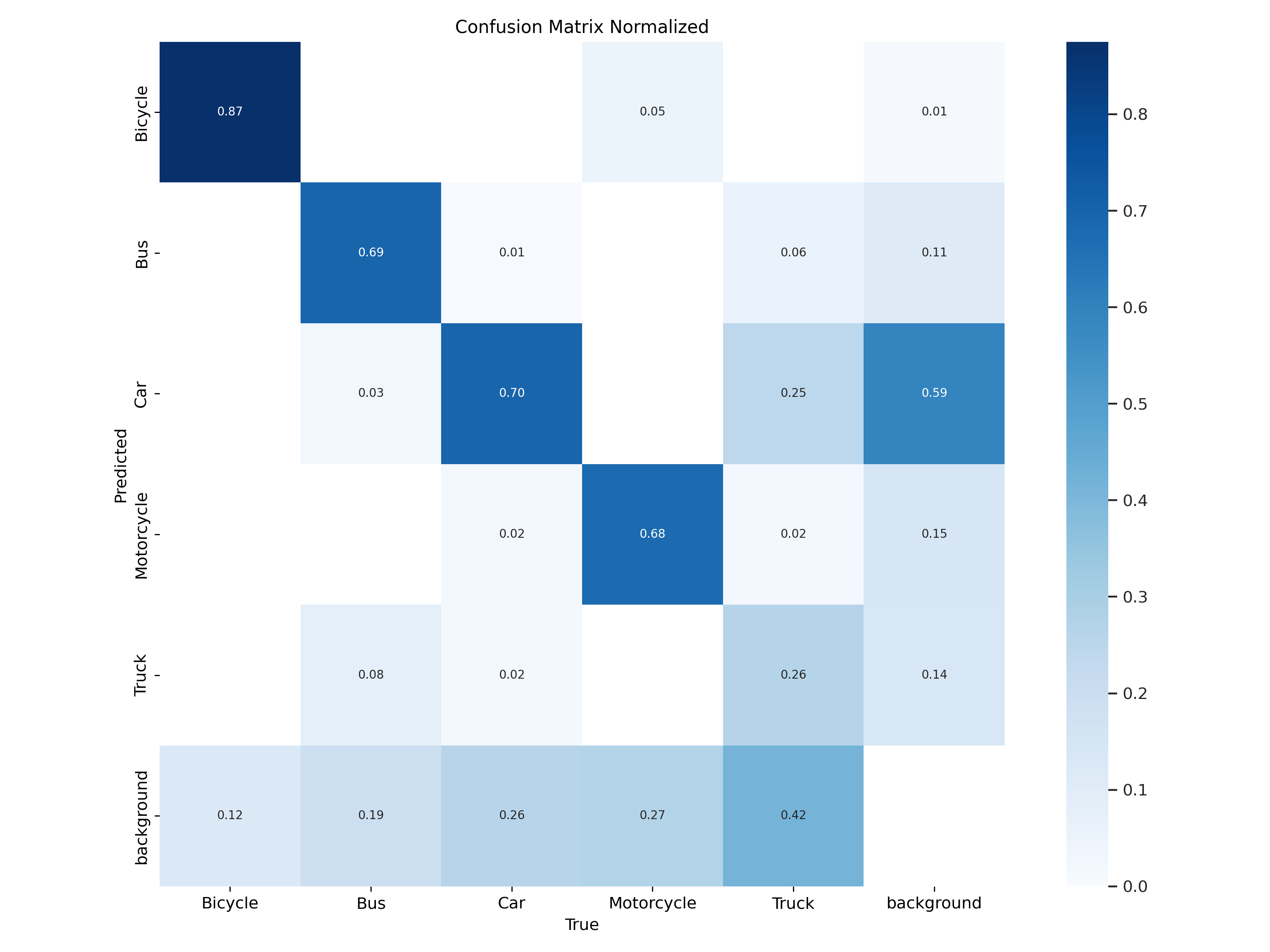}
\end{adjustwidth}
\caption{YOLOv5l6s normalized confusion matrix.}
\label{Figure:10}
\end{figure}

The confusion matrix for YOLOv5m6s reveals a moderate performance across all classes. The model shows a balanced ability to classify objects. However, like YOLOv5s6s, there is some confusion between Bus and Truck and Motorcycle and Bicycle, indicating that the model's performance may be limited by the visual similarity between certain vehicle classes (Figure \ref{Figure:9}).YOLOv5l6s exhibits a strong overall performance, with higher values compared to other variants, particularly in the Car class. However, there remains a struggle in recognizing the Motorcycle and Bicycle categories, although it is less pronounced than in YOLOv5s6s and YOLOv5m6s. The larger model size and increased complexity might contribute to its improved accuracy (Figure \ref{Figure:10}).

\begin{figure}[H]
\begin{adjustwidth}{-\extralength}{0cm}
\centering
\includegraphics[height=15cm]{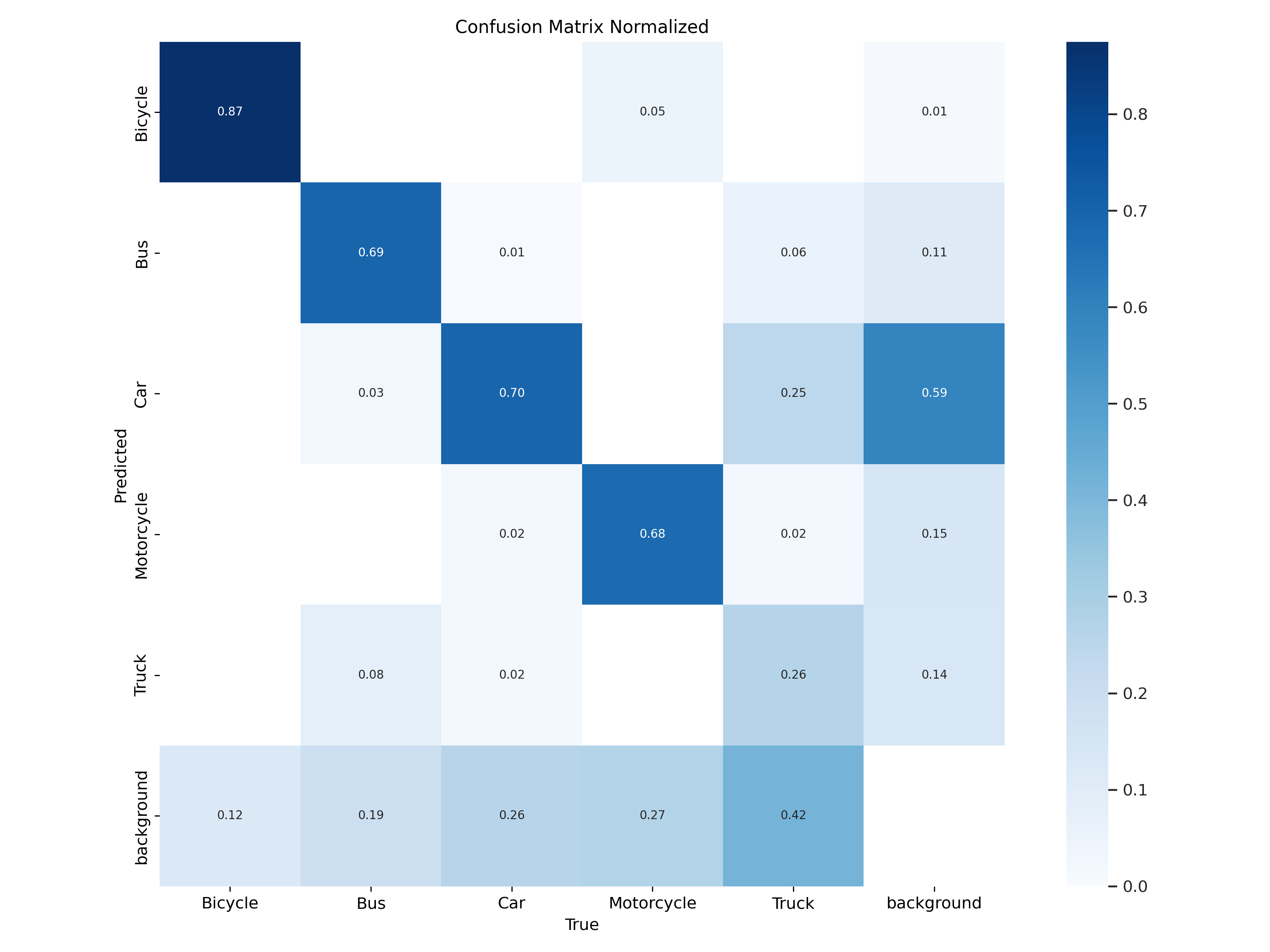}
\end{adjustwidth}
\caption{YOLOv5x6s normalized confusion matrix.}
\label{Figure:11}
\end{figure}

The confusion matrix for YOLOv6x6s shows high accuracy in the Bicycle and Background categories. However, the model demonstrates more confusion in classifying Truck and Bus, similar to other variants. Despite this, the YOLOv6x6s variant performs better in distinguishing between classes, Motorcycle and Car (Figure \ref{Figure:11}).Overall, the confusion matrices highlight that while all models perform well on Car, there is consistent difficulty across variants in distinguishing between Bus and Truck and Motorcycle and Bicycle; most of the classes are not identified correctly, rather identified as background. YOLOv5l6s generally offers the best performance across all categories, likely due to its larger model size

\section{Discussion}
The performance of all the YOLOv5 versions for vehicle detection, including five classes, has been compared in this study. The evaluation of YOLOv5 variants over 40 epochs reveals unique patterns in their precision and recall. Early in the training process, precision fluctuates notably across all variants, especially for YOLOv5s6s and YOLOv5m6s, suggesting that the models initially did not perform well. As the training progresses, these fluctuations diminish, leading to more stable precision levels.

Recall shows a gradual and consistent improvement across all YOLOv5 variants, with YOLOv5s6s and YOLOv5m6s showing the best performances. These models demonstrate a marked ability to improve their detection of correct objects. However, YOLOv5n6s and YOLOv5l6s exhibit a slower increase in recall, indicating that while they are effective, they may be less capable of detecting all relevant objects.

The F1-score, which balances precision and recall, provides further insight into the overall performance of the models. YOLOv5n6s reaches the highest peak F1 score. However, all variants struggle with specific classes, especially Motorcycle, where the F1 scores are lower.  The mAP, especially mAP50 and mAP50-95, demonstrates that while all models improve over time, YOLOv5s6s, YOLOv5m6s, and YOLOv5x6s show more consistent and robust performance, suggesting their superior adaptability and generalization. Furthermore, the confusion matrices support these findings, showing that while all variants perform well in classifying Car, there is persistent difficulty in distinguishing between Bus and Truck as well as Motorcycle and Bicycle.

In summary, while the models show a significant increase in precision and recall as training progresses, challenges remain in identifying certain classes, particularly confusion between visually similar objects. These findings suggest that further refinement in training strategies could enhance the performance of YOLOv5 variant models in real-world environment.

\section{Conclusions}
The performance of the YOLOv5 variants has been thoroughly evaluated in this study, focusing on their ability to detect various vehicle classes, including Bicycle, Bus, Car, Motorcycle, and Truck. The findings reveal that YOLOv5n6s excels in precision and recall, particularly in classifying Car, whereas YOLOv5s6s and YOLOv5m6s show improvements in recall, making them more better at identifying all relevant objects. However, all variants consistently struggle with distinguishing between similar classes such as Bus and Truck and Motorcycle and Bicycle, indicating areas where the models could be further refined.

Future work could explore enhancing these models' performance by focusing on improving the detection of challenging classes, mainly Background. Additional studies could involve expanding the evaluation to different object detection tasks~\cite{hussain2023review} or incorporating more diverse and complex datasets~\cite{aydin2023domain}, potentially including occluded objects~\cite{hussain2023child}. The research could also extend to applying these YOLOv5 variants to other domains where object detection plays a critical role, such as autonomous driving, robotics, and renewable energy systems, thereby broadening the applicability and impact of these models.

\begin{adjustwidth}{-\extralength}{0cm}

\bibliographystyle{unsrt}  
\bibliography{ref}  

\end{adjustwidth}
\end{document}